\documentclass[11pt,a4paper]{article}
\usepackage[hyperref]{emnlp2018}
\usepackage{times}
\usepackage{latexsym}
\usepackage{amsmath}
\usepackage{multirow}
\usepackage{amssymb,amsfonts}
\usepackage{arydshln}
\usepackage{amsmath}
\usepackage{rotating}
\usepackage{color}
\usepackage{subfigure}
\usepackage[small]{caption}
\usepackage[]{microtype}
\usepackage[subtle]{savetrees}

\usepackage{adjustbox}

\usepackage{url}

\newcommand{\ptheta}{p_{{\boldsymbol \theta}}}

\newcounter{notecounter}
\newcommand{\enoteson}{\long\gdef\enote##1##2{{
\stepcounter{notecounter}
\large\bf
\hspace{1cm}\arabic{notecounter} $<<<$ ##1: ##2
$>>>$\hspace{1cm}}}}
\newcommand{\enotesoff}{\long\gdef\enote##1##2{}}

\usepackage{color}
\usepackage{xcolor}
\usepackage[disable]{todonotes}   
\usepackage{blindtext}

\aclfinalcopy 

\newcommand{\note}[4][]{\todo[author=#2,color=#3,size=\scriptsize,fancyline,caption={},#1]{#4}} 
\newcommand{\katha}[2][]{\note[#1]{Katharina}{yellow!40}{#2}}   
   
\enoteson
\enotesoff

\title{Neural Transductive Learning and Beyond:\\Morphological Generation in the Minimal-Resource Setting}

\author{  
  Katharina Kann\\ 
  Center for Data Science\\
  New York University, USA\\
  \texttt{kann@nyu.edu} 
  \\\And
  Hinrich Sch\"{u}tze \\
  CIS\\
  LMU Munich, Germany\\
  \texttt{inquiries@cislmu.org}
}
\date{}

\def\figref#1{Figure~\ref{fig:#1}}
\def\figlabel#1{\label{fig:#1}\label{p:#1}}

\def\tabref#1{Table~\ref{tab:#1}}
\def\tablabel#1{\label{tab:#1}\label{p:#1}}

\def\secref#1{\S\ref{sec:#1}}
\def\seclabel#1{\label{sec:#1}\label{p:#1}}
\def\eqref#1{Eq.~\ref{eqn:#1}}

\begin{document}
\maketitle
\begin{abstract}
Neural state-of-the-art sequence-to-sequence (seq2seq) models
often do not perform well for small
training sets.
We address paradigm completion, the
morphological task of, given a partial paradigm, 
generating all
missing forms.
We propose two new methods
for 
the minimal-resource setting:  (i)
\emph{Paradigm transduction}:  Since we assume
only few paradigms available for training,
neural seq2seq models are able to capture relationships between paradigm
cells,
but
are tied to the idiosyncracies of the training set.
Paradigm transduction mitigates this problem by
exploiting the input subset of inflected forms at test time. 
(ii) \emph{Source selection with high precision (SHIP)}:
Multi-source models which learn to automatically select one or 
multiple sources to predict a target inflection do not
perform well in the minimal-resource
setting. 
SHIP is an alternative to
identify a reliable
source if training data is limited.
On a 52-language benchmark dataset, 
we outperform the previous state of the art by
up to $9.71\%$ absolute accuracy.
\end{abstract}

\section{Introduction}
\seclabel{intro}
Morphological generation of previously
unencountered word forms is a crucial problem
in many areas of natural language processing
(NLP). High performance
can lead to better systems for
downstream tasks, e.g., machine translation \cite{tamchyna-wellerdimarco-fraser:2017:WMT}.
Since existing lexicons have limited coverage,
learning morphological
inflection patterns from labeled data is
an important mission and has recently been the subject of multiple shared tasks \cite{cotterell-sigmorphon2016,cotterell-conll-sigmorphon2017}.

In morphologically rich languages, words inflect, i.e., they change their surface form in oder to express certain properties, e.g., number or tense.
A word's canonical form, which can be found in a dictionary, is called the lemma, and the set of all inflected forms is referred to as the lemma's paradigm.
In this work, we address paradigm completion (PC), the
morphological 
task of, given
a partial paradigm of a lemma,
generating all of its missing forms.
For 
the partial paradigm represented by the input subset \{(``Schneemannes'', GEN;SG),
(``Schneem\"{a}nnern'', DAT;PL)\}
of the German noun
``Schneemann'' shown in \figref{snowman},
the goal of
PC is to
generate 
the output subset consisting of the six remaining forms.

\begin{figure}[t]
\centering
\includegraphics[width=0.92\columnwidth]{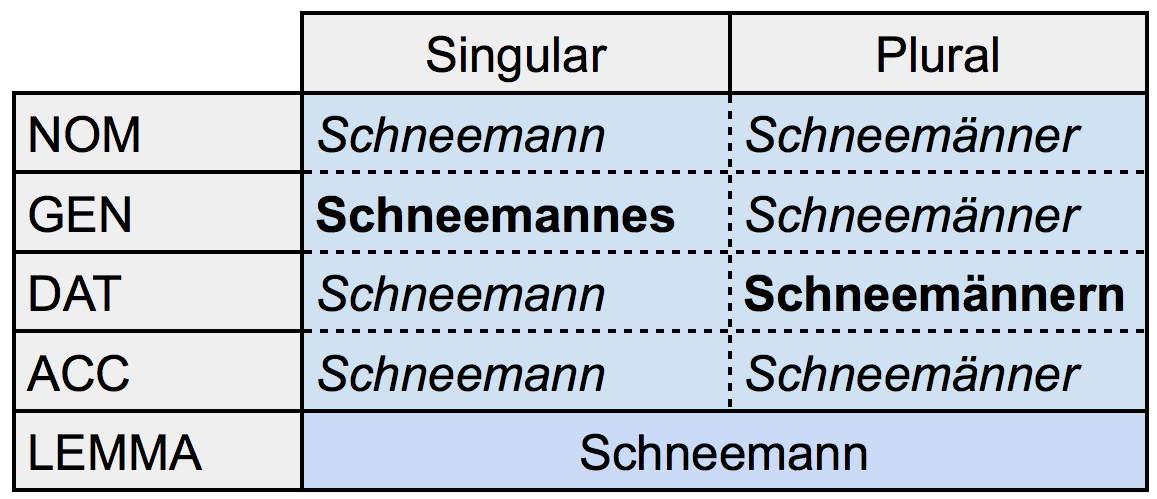}
\caption{The paradigm of the German  noun
    ``Schneemann'' (``snowman''). In this running
    example, the input subset is
    bold, the output subset italic.}
\figlabel{snowman}
\end{figure}
Neural seq2seq models define the
state of the art for 
morphological generation if training sets are  large;
however, 
they have been less
successful in the low-resource setting \cite{cotterell-conll-sigmorphon2017}. In this paper, we
address an even more extreme 
\emph{minimal-resource} setting: 
for some of our experiments, 
our
training sets only contain $k \approx 10$ paradigms. 
Each paradigm has multiple cells, so the number
of \emph{forms} (as opposed to the number of \emph{paradigms}) is not
necessarily minimal.
However, we will see that
generalizing from paradigm to paradigm is a key challenge,
making 
the number of paradigms
a good measure of the effective training set size.

We propose two PC methods for the minimal-resource
setting: \emph{paradigm transduction} and \emph{source selection with high precision (SHIP)}.  We
define a learning algorithm as \emph{transductive}\footnote{In order to avoid ambiguity, ``transduction" is never used in the sense of string-to-string transduction in this paper.} if its
goal is to generalize from specific training examples to
specific test examples \cite{vapnik98theory}. In contrast, inductive
inference learns a general model
that is independent of any test set. Predictions of
transductive inference for the same item are different for different test
sets. There is no such
dependence in inductive inference.
Our motivation for transduction
is that, in the minimal-resource setting,
neural seq2seq models capture
relationships between paradigm cells like affix substitution
and umlauting, but  are tied to the
idiosyncracies of the $k$ training paradigms. 
For example, if all source forms in the training set start with ``b'' or ``d'', a purely inductive model may then be unable to generate targets with
different initials.
By  transductive inference
on the information available in the input subset at test time, i.e., the given partial paradigm,
our model can learn
idiosyncracies. For example, if the input subset sources 
start with ``p'', we can learn to generate
output subset targets that start with ``p''. Thus, 
we exploit the input subset for learning idiosyncracies at test
time  and then generate the output subset 
using a modified model.  This
setup
employs standard inductive training (on the training set) for
learning general rules of inflectional
morphology
and transductive inference (on the test set) for learning
idiosyncracies.
Our use of transduction is innovative in that
most previous work has addressed unstructured problems
whereas our problem is structured: we complete a
paradigm, a complex structure of forms, each of them labeled
with a morphological tag. Thus, the test set contains labels,
whereas, in transduction for unstructured problems, the
test set is a flat set of unlabeled instances. We view
our work as an extension of transduction to the structured
case,
even though
not all elements of the theory developed by
\newcite{vapnik98theory} 
carry over.

The motivation for our second PC
method for limited training data,
SHIP, is as follows.
Multi-source models can learn which combination of sources
most reliably predicts the target in the high-resource,
but less well in the minimal-resource setting. 
SHIP models the relationship between paradigm slots
using
edit trees \cite{chrupala2008towards}, in
order to measure how deterministic each transformation 
is. Then, it identifies the 
most deterministic source slot for the generation of each target inflection.

Paradigm transduction and SHIP can be employed separately or in combination. 
Our experiments show that, in an extreme minimal-resource setting, a combination of SHIP
and a non-neural approach is most effective;
for slightly more data, a combination of a neural model, paradigm transduction and SHIP obtains the best results.

\paragraph{Contributions.}
(i) We introduce neural paradigm transduction, which
exploits the
structure of the PC task to mitigate the negative effect of limited training data.
(ii) We propose SHIP, a new algorithm for
picking a single
reliable source for PC in the minimal-resource setting.
(iii) On average over all languages of a 52-language
benchmark dataset,
our approaches 
outperform state-of-the-art baselines
by up to $9.71\%$ absolute accuracy.

\section{Paradigm Completion}
In this section, we formally define our
task, developing the notation for the rest of the paper.

Given the set of morphological tags $T(w)$ of a lemma $w$, we define the paradigm of $w$ as the set of tuples of inflected form $f_k$ and tag $t_k$:
\begin{equation}
  \pi(w) = \big\{ \big( f_k[w], t_{k} \big) \big\}_{t_k \in T(w)}
\end{equation}
The example in \figref{snowman} thus corresponds to: 
$\pi$(Schneemann) =
$\big\{\big($``Schnee\-mann'',
NOM;SG$\big)$ \ldots
$\big($``Schneem\"{a}nner'', ACC;PL$\big)\big\}$.

A training set in our setup consists of complete paradigms, i.e., all
inflected forms of each lemma are available. This simulates
a setting in which a linguist
annotates complete
paradigms, as done, e.g., in \newcite{sylak2016remote}.
In contrast, each element of the test set is a partial
paradigm, which we refer to as
the \emph{input subset}.
This simulates a setting
in which we collect all forms of a lemma occurring in a
(manually or automatically) annotated
input corpus; this set will generally not be complete. 
The
PC task consists of generating  the
\emph{output subset} of the paradigm, i.e., the forms 
belonging to form-tag pairs which are missing from the collected subset.

\section{Method}
\seclabel{method}
Our approach for PC is based on MED (\textit{Morphological Encoder-Decoder}), a state-of-the-art model for morphological generation in the high-resource case, which was
developed by \newcite{kann-schutze:2016:P16-2}.
In this section, we first cover required background on MED and then introduce our new approaches.

\subsection{MED}
\label{subsec:med}
\paragraph{Input and output format.} 
MED converts one inflected form of a paradigm into another, given the two respective tags.
Thus, the input of MED is a sequence of subtags of the source and the target form
(e.g., NOM and SG are subtags of NOM;SG), as well as the characters of 
the source form. All elements are represented by embeddings, which are trained together with the model.
The output of MED is the character sequence of the target inflected form.

An example from the paradigm in \figref{snowman} is:

\begin{footnotesize}
\vspace{0.3cm}
\ \hspace{-0.4cm}\begin{tabular}{l}
\textbf{INPUT: } {DAT}$^S$ {PL}$^S$ {GEN}$^T$ {SG}$^T$ S c h n e e m \"{a} n n e r n \\
\textbf{OUTPUT: } S c h n e e m a n n e s 
\end{tabular}
\vspace{0.3cm}
\end{footnotesize}

\paragraph{Encoder.} 
The model's encoder consists of a bidirectional gated recurrent neural network (GRU)
with a single hidden layer.
It reads an input vector sequence 
$x = (x_1, \ldots, x_{X_t})$
and encodes it from two opposite directions into two hidden representations
$\overrightarrow{h_t}$ and $\overleftarrow{h_t}$ as
\begin{align}
  \overrightarrow{h_t} &= \text{GRU}(x_t, \overrightarrow{h_{t-1}})\\
  \overleftarrow{h_t} &= \text{GRU}(x_t, \overleftarrow{h_{t+1}})
\end{align}
which are concatenated to
\begin{equation}
  h_t = \left[\overrightarrow{h_t}; \overleftarrow{h_t}\right]
\end{equation}

\paragraph{Decoder.}
The decoder, another GRU with a single hidden layer, defines a probability distribution over
the output vocabulary, which, for paradigm completion, consists of the characters in the language, as
\begin{align}
  p(y) = \prod_{t=1}^{T_y} \text{GRU}(y_{t-1}, s_{t-1}, c_t)
\end{align}
$s_t$ denotes the state of the decoder at step $t$, and $c_t$ is the sum of the hidden representations of the encoder, 
weighted by an attention mechanism.

Additional background on the general model architecture is given in \newcite{bahdanau2015neural}; 
details on MED can be found in \newcite{kann-schutze:2016:P16-2}.

\subsection{Semi-supervised MED}
\label{subsec:semisupmed}
In order to make use of unlabeled data with MED, \newcite{kann2017unlabeled} defined an auxiliary autoencoding task and proposed a multi-task learning approach.

For this extension, an additional symbol is added to the input vocabulary. 
Each input is then of the form $\left(\text{\textbf{A}} \mid
M^+\right)\Sigma^+$, with
\textbf{A} being a novel tag for autoencoding,
$\Sigma$ being the alphabet of the language, and
$M$ being
the set of morphological subtags of the source and the target. 
As for the basic MED model, all parts of the input are represented by embeddings.

The training objective is
to maximize the joint likelihood for the tasks of paradigm completion and autoencoding:
\begin{eqnarray*}
  {\cal L}({\boldsymbol \theta}) = & \sum_{(s, t^S, t^T, w)
    \in {\cal D}} &\log \ptheta (w \mid e_\theta(t^S, t^T, s) ) \\ 
  & + \sum_{a \in {\cal A}} & \log \ptheta (a \mid e_\theta(a))
\end{eqnarray*}
where  $\cal A$ is a 
set of autoencoding examples, $e_\theta$ is the encoder, and
$\cal D$ is a labeled training set  of
tuples of source $s$, morphological source tag $t^S$, morphological target tag $t^T$, and  target $w$.

\subsection{MED for Paradigm Completion}
\label{subsec:training_data}
MED was originally developed for morphological reinflection.
Thus, it operates on pairs consisting of a single source and a single target form.
In order to use it for paradigm completion, where multiple source forms 
are given, and multiple target forms are expected, we
convert the given data into a suitable format in the way described in the following.

For a lemma $w$, let $J(w)$ be the set of tags in the input
subset. Recall that $J(w)$ is a subset of 
$T(w)$, the set of all tags, at test time, but that training
paradigms are complete, i.e., $J(w)=T(w)$ for the training set.

For both training of the inductive model and 
paradigm transduction, we generate $|J(w)|(|J(w)|-1)$ training examples
\[
 (t_i, t_j, f_i[w]) \mapsto f_j[w]
\]
one for
each pair of different tags in $J(w)$.
We also generate autoencoding training examples
for all tags in $J(w)$
(removing duplicates):
\[
  \big({\tt \textbf{A}}, f_i[w] \big) \mapsto f_i[w] 
\]

{For the German lemma ``Schneemann'', assume:}

\begin{tabular}{l}
\centering
$J(\mbox{Schneemann})=\{\mbox{GEN;SG}, \mbox{DAT;PL}\}$
\end{tabular}

{at test time. 
We then produce the following training examples for paradigm
transduction:}\\[0.05cm]

\begin{footnotesize}
\hspace{-0.6cm}
\begin{tabular}{l}
({DAT}$^S$ {PL}$^S$ {GEN}$^T$ {SG}$^T$ Schneem\"{a}nnern) $\mapsto$ Schneemannes  \\
({GEN}$^S$ {SG}$^S$ {DAT}$^T$ {PL}$^T$ Schneemannes) $\mapsto$ Schneem\"{a}nnern\\
(\textbf{A} Schneemannes) $\mapsto$ Schneemannes\\
(\textbf{A} Schneem\"{a}nnern) $\mapsto$ Schneem\"{a}nnern
\end{tabular}
\end{footnotesize}

For completing 
a partial paradigm, we then select one source form per target slot
(the lemma, unless stated otherwise) 
and create all forms corresponding to the tags in $J(w)\setminus T(w)$
one by one.

\begin{figure*}[!t]
\centering
\includegraphics[width=\textwidth]{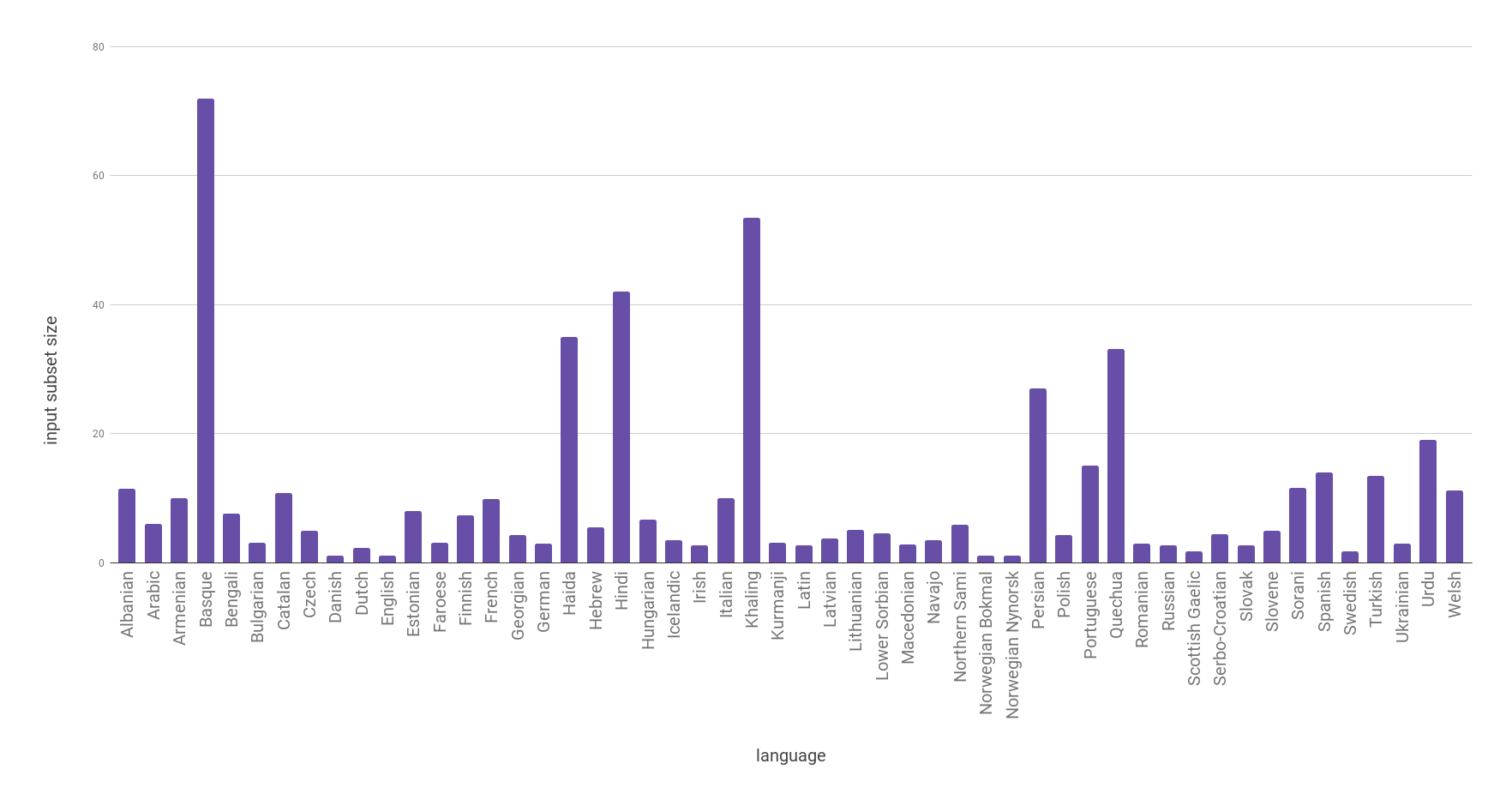}
\caption{Average amount of sources in the input subset for paradigm transduction, per language.}
\figlabel{amount_ft}
\end{figure*}
\subsection{Paradigm Transduction}
\seclabel{pt}
\paragraph{Motivation.}
In the minimal-resource setting, 
parameter estimates  are tied to the idiosyncracies
of the lemmas seen in training, due to overfitting. Our
example in
\secref{intro} is that the model has difficulties producing
initial letters not seen during training.
However, within each paradigm, forms are generally similar;
thus, input subset sources contain valuable information
about how to generate output subset targets.  Based on
this observation, we solve the problem of overfitting by
transduction: we teach the model test idiosyncracies
by training it on the input subset before generating the output
subset.

\paragraph{Method description.}
We first train a general model
on the  training set in the standard supervised learning
setup, i.e., the setup which is called inductive inference by
\newcite{vapnik98theory}.
At test time,
we take the general model as initialization and continue training on examples generated from 
the input subset as described in \S\ref{subsec:training_data}.
We do this  separately for each lemma,
satisfying the defining criterion of transductive
inference that 
predictions depend on the test
data. Also,  different
input subsets (i.e., different subsets of the same paradigm)
can in general make
different predictions on an output subset target.

Paradigm transduction is expected to perform best in a setting in which 
many forms of each paradigm are given as input, i.e., when
$|J(w)|$ is big.
In \figref{amount_ft} we show the average sizes of the input subsets
for all languages in our experiments.

\begin{figure}[]
  \centering
  \includegraphics[width=0.74\columnwidth]{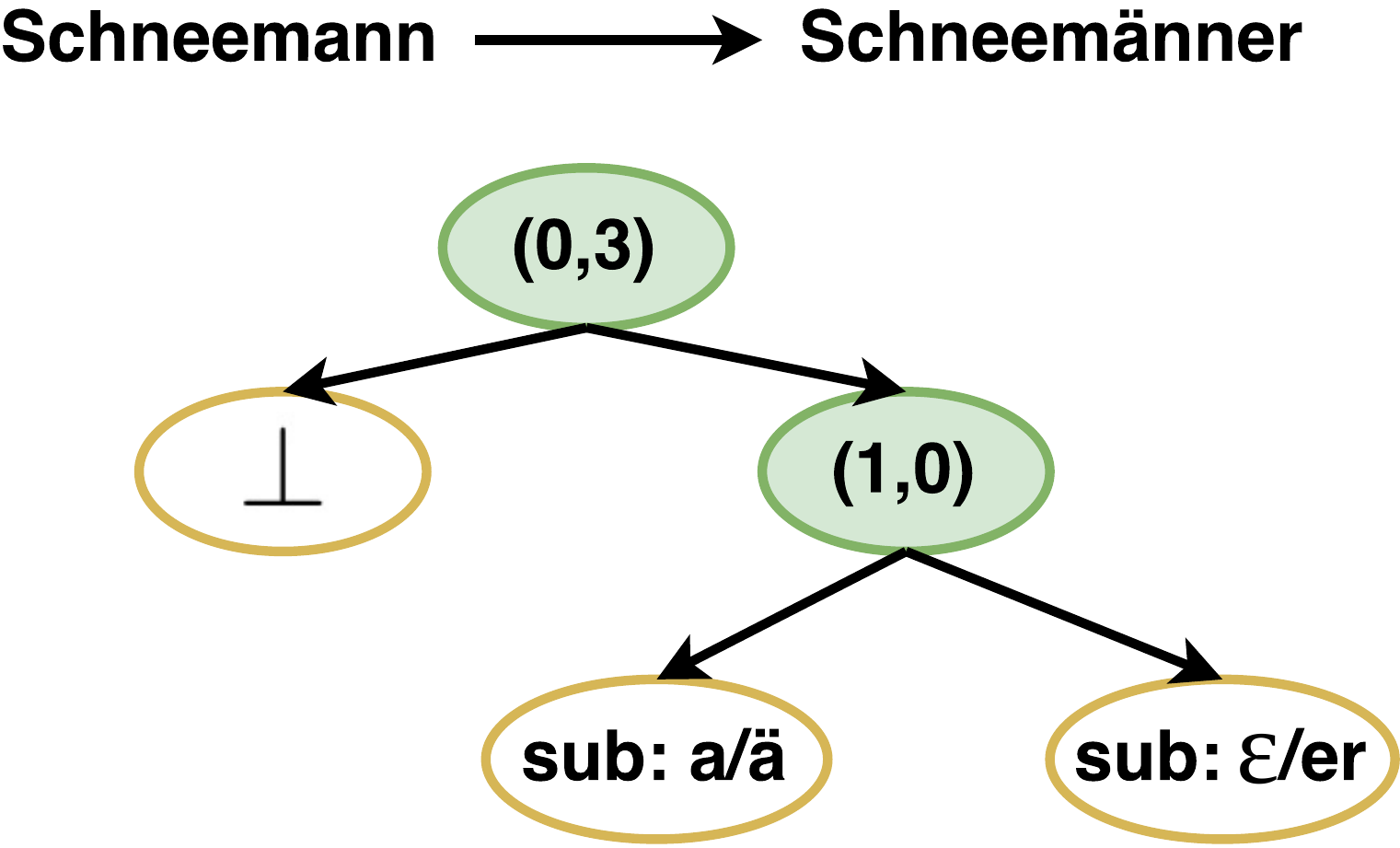}
  \caption{Edit tree example.
Each node
    gives lengths of the parts before/after 
    LCS, e.g.,
the root has LCS ``Schneem'', before part
$\epsilon$ and after part ``ann'', thus the lengths are
    ``(0,3)''.
    ``sub'' = ``substitution''.}
  \label{fig:edittree}
\end{figure}
\begin{figure*}[t]
  \centering
  \includegraphics[width=0.7\columnwidth]{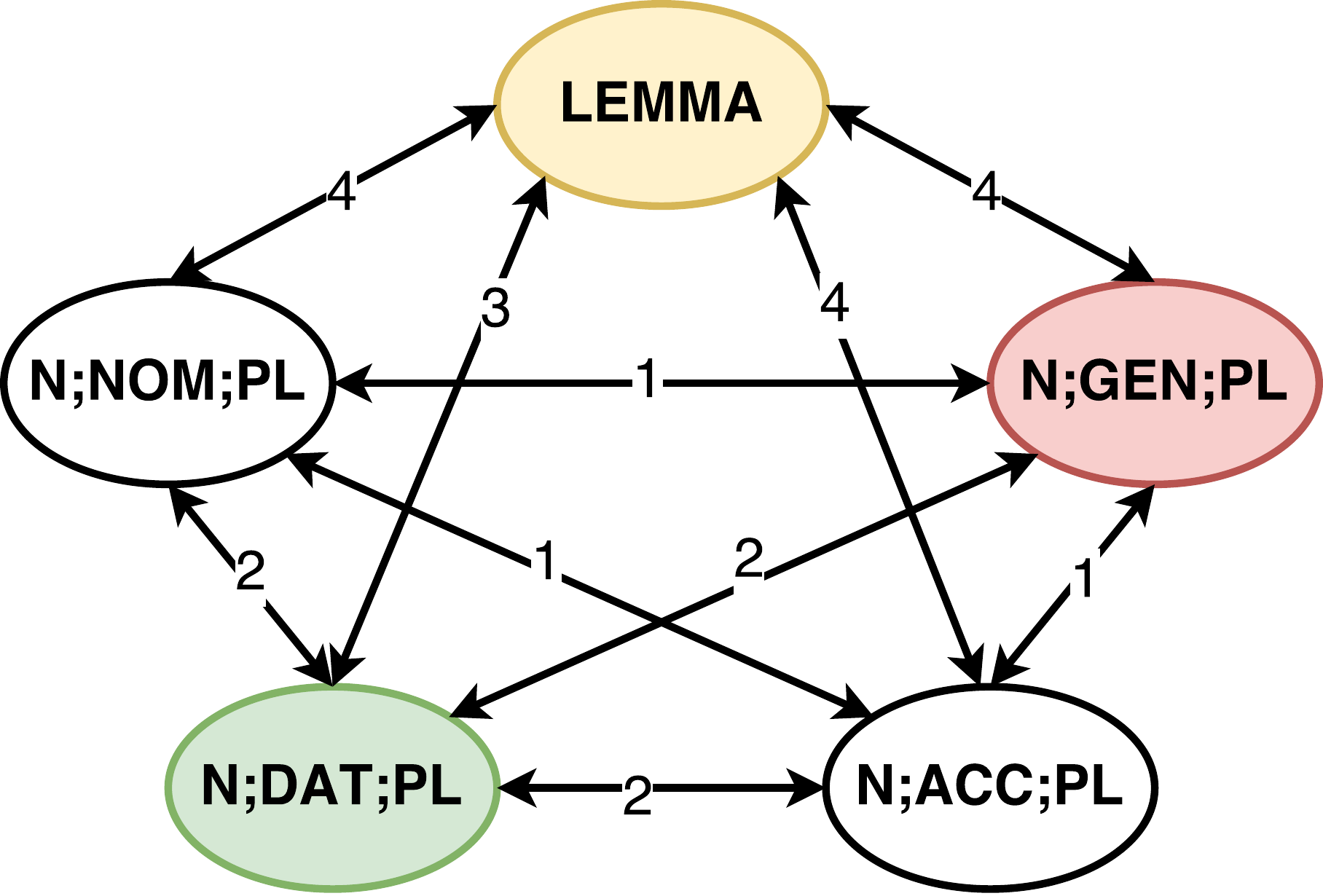}
  \hspace{1.3cm}
  \includegraphics[width=0.7\columnwidth]{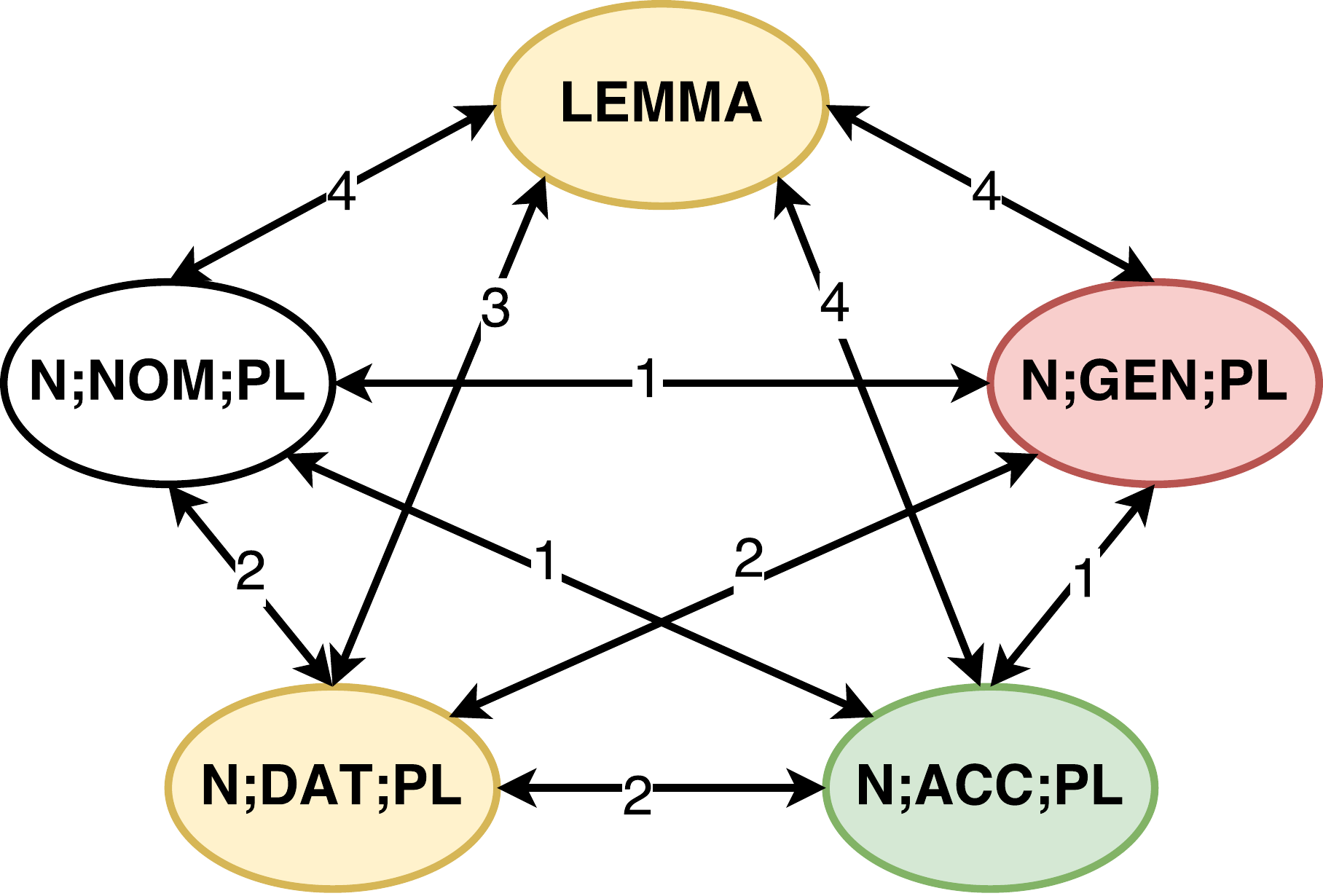}
\caption{SHIP example for German plural forms (SET1). 
For the graph constructed in training (see \secref{ship}),
subgraphs are extracted in testing for input subset sizes
two (left) and three (right). Input subset: yellow and
green. Output subset: white and red.
For generation of the target shown in red, SHIP selects the
source shown in green.\figlabel{shipexample}}
\end{figure*}
\subsection{Source Selection with High Precision}
\seclabel{ship}
During PC,
some sources contain more
information relevant to generating certain targets than
others. For instance, the nominative singular and accusative
singular in German are generally identical
(cf.\ \figref{snowman}); thus, for generating the accusative singular,
we should use the nominative singular as
source if it is available---rather than, say, the dative
plural.

In fact, for many languages, the entire paradigm
of most lemmas
is
deterministic 
if the right source forms are known and used for the right targets.
A set of forms that determines all other
inflected forms is called \emph{principal parts}
\cite{finkelstump2007}. Based on this theory,
\newcite{cotterell-sylakglassman-kirov:2017:EACLshort} 
induce
topologies and jointly decode entire paradigms, thus
making use of all available forms. However, their method is
only applicable if 
 good estimates of the probabilities
$p(f_j[w] | f_i[w])$ for source $f_i[w]$ and
target $f_j[w]$ can be obtained, 
and they train on hundreds of paradigms
per part of speech (POS) and language, which are not available in our setup.

We propose an alternative for the minimal-resource setting:
SHIP, which
selects a single best source
for each target and is
based on edit trees.
An edit tree $e(f_i[w], f_j[w])$ is  a transformation from a
source  $f_i[w]$ to a target $f_j[w]$
\cite{chrupala2008towards}; see
\figref{edittree}.
It is constructed by first
determining the longest common substring (LCS)
\cite{gusfield1997algorithms} of $f_i[w]$ and $f_j[w]$ and then
modeling the prefix and suffix pairs of the LCS recursively.
In the case of an empty LCS, $e(f_i[w],f_j[w])$ is the
substitution operation that replaces $f_i[w]$ with $f_j[w]$.

We construct edit trees for each pair $(f_i[w], f_j[w])$
in the training set, count the number $n_{ij}$ of
different edit trees  for 
$t_i \mapsto t_j$, and construct a fully connected
graph.
The tags are nodes of the graph, and the counts $n_{ij}$ are weights.
Edges are undirected, since edit trees are bijections 
(cf. \figref{shipexample}). 
We then interpret the weight of an edge as a measure of
the (un)reliability of
the corresponding two source-target relationships.
Our intuition  is that the fewer different edit trees
relate source and target, the more
reliable the source is for generating the target.

At test time, we find for each target $t_j$ 
 a source  $t_k$ such that $n_{kj} \leq n_{ij} \forall i
 \in J(w)$.   We then use $f_k[w]$ to
generate $f_j[w]$. Again, \figref{shipexample}
shows examples.

\section{Experiments}
\seclabel{experiments}

\subsection{Data}
\label{subsec:data}

We run experiments on the datasets from task
2 of the CoNLL--SIGMORPHON 2017 shared task, which have been 
created using UniMorph \cite{kirov2018unimorph}.
We give a short overview here;
see 
\cite{cotterell-conll-sigmorphon2017}
for details.
The dataset contains, for each of 52 languages,
a development set of 50 partial paradigms, a test
set of
 50 partial paradigms,
and three training sets of 
complete paradigms.
Training set sizes are
10 (SET1), 
50 (SET2), and  200  (SET3).
Recall that
we view 
the number of
paradigms (not the number of forms) as the best
measure of the amount of training data available.
Even for SET3, there are 
only 200 lemmas per language in the training set, which are additionally 
distributed over
multiple POS tags, compared to $>$600
lemmas per POS used by
\newcite{cotterell-sylakglassman-kirov:2017:EACLshort}.
We, thus, want to emphasize that 
all settings---SET1, SET2, and SET3---can be considered
low-resource. 

We produce training sets for our encoder-decoder as
described in
\S\ref{subsec:training_data}, but limit the total
number of training examples to 200,000.

\subsection{Hyperparameters}
With our hyperparameters, we follow
\newcite{kann-schutze:2016:SIGMORPHON}.
In particular, 
our encoder and decoder GRUs have 100-dimensional hidden states. Our
embeddings are 300-dimensional. 
For training, we use stochastic gradient descent, ADADELTA \cite{zeiler2012adadelta}, and 
minibatches of size $20$.
After experiments on the development set, we decide on training 
SET1, SET2, and SET3 models for 50, 30, and 20 epochs, respectively.
For paradigm transduction, we train all models for 25 additional epochs.

\subsection{Baselines}
In the following, we describe our baselines.
COPY, MED, and PT are used for ablation
and SIG17 for comparison with the state of
the art.

\paragraph{COPY.} As targets in many paradigm cells in many
languages are identical to the
lemma, we consider a copy baseline that  simply
copies the lemma.

\paragraph{MED.} This
is 
the model by \newcite{kann-schutze:2016:P16-2}, which performed best
at SIGMORPHON 2016.
For decoding, the lemma is
used. Since MED is designed for the high-resource
setting,  we 
do not expect good
performance
for our minimal-resource scenario, but the comparison
shows how much our
enhancements
improve performance.

\paragraph{Pure paradigm transduction (PT).} PT is
a seq2seq
model exclusively trained on the input subset.
Its performance sheds light on the importance of the initial 
inductive training.

\paragraph{SIG17. }
SIG17 is the
official baseline of 
the CoNLL--SIGMORPHON 2017 shared task,
which was developed to perform well with very little
training data.
Its design follows \newcite{liu-mao:2016:SIGMORPHON}:
SIG17 first
aligns each input lemma and output inflected form.
Afterwards,
it assumes that each aligned pair
can be split into a prefix, a stem, and a suffix.
Based on this alignment, the system extracts prefix (resp.\ suffix)
rules from the prefix (resp.\ suffix) pairings.
At test time, 
suitable rules are applied to the input string to
generate the target; more details can be found in 
\newcite{cotterell-conll-sigmorphon2017}.

\begin{table}[!t]
  \centering
    \setlength{\tabcolsep}{5.5pt} 
  \begin{tabular}{l | c c c }
   & SET1 & SET2 & SET3\\\hline
   \textit{BL}: COPY & .0810 & .0810 & .0810 \\
   \textit{BL}: MED & .0004 & .0432 & .4211 \\
   \textit{BL}: PT & .0833 & .0833 & .0775 \\
   \textit{BL}: SIG17 & .5012 & .6576 & .7707 \\
   SIG17+SHIP & \textbf{.5971} & .7355 & .8008 \\
   MED+PT & .5808 & .7486 & .8454 \\
   MED+PT+SHIP & .5793 & \textbf{.7547} & \textbf{.8483} \\
  \end{tabular}
  \caption{Accuracy on PC 
  for SIG17+SHIP
  (the shared task baseline SIG17 with  SHIP), MED+PT (MED with paradigm transduction), MED+PT+SHIP (MED with paradigm transduction and
  SHIP), as well as all baselines (\textit{BL}). Results are averaged over all languages, and best results are in bold; detailed accuracies for all languages can be found in Appendix \ref{app:C}.
  \tablabel{small_results}}
\end{table}

\begin{table*}[h]
  \centering
\setlength{\tabcolsep}{5.pt}
  \begin{tabular}{l |lll}
  \multicolumn{1}{c|}{\bf input}  & \multicolumn{3}{c}{\bf output} \\
                & MED & PT & MED+PT \\\hline
  Schneemann N;GEN;PL &  GetG\"{a}chen &  Scnneeeeennnnnnnnnnnnnnnnnnnnn & Schneem\"{a}nner \\
  dish V;V.PTCP;PRS & dising &  dish &  dishing \\
  creer V;SBJV;PRS;1;PL & crezcamos &  creyemos &  creamos
  \end{tabular}
  \caption{Analysis of the outputs of MED, PT, and MED+PT 
    for SET2. Top to bottom: German, English, Spanish. MED and PT produce incorrect, MED+PT
    correct inflections.\tablabel{erroranalysis}}
\end{table*} 
\subsection{Results}
Our results are shown in 
\tabref{small_results}.
For SET1, SIG17+SHIP obtains the highest accuracy, while,
for SET2 and SET3, 
MED+PT+SHIP performs best. This difference can be easily explained by the fact that
the performance of neural networks decreases rapidly for smaller training sets, and, while paradigm transduction
strongly
mitigates this problem, it cannot completely eliminate it.
Overall, however, SIG17+SHIP, MED+PT, and MED+PT+SHIP all outperform the baselines by a wide margin
for all settings.

\paragraph{Effect of paradigm transduction.}
On average, MED+PT clearly outperforms SIG17, the strongest baseline: by
.0796 (.5808-.5012) on SET1, .0910 (.7486-.6576) on SET2,
and .0747 (.8454-.7707) on SET3.

However, 
looking at each language individually (refer to Appendix \ref{app:C} for those results), we find that 
MED+PT performs poorly for a few languages, namely Danish,
English, and Norwegian (Bokm\aa l \& Nynorsk).
We hypothesize that this can most likely be explained by the
size of the input subset of those languages being small
(cf. \figref{amount_ft} for average input subset sizes per language).
Recall that the input subset
is explored by the model during transduction.
Most poorly performing
languages have input subsets containing only
the lemma;
 in this case
paradigm transduction reduces to
autoencoding the lemma.  Thus, we conclude that paradigm transduction can 
only improve over MED if two or more sources
are given.

Conversely, 
if we consider only the languages with an average input subset size of
more than $15$ (Basque, Haida, Hindi, Khaling, Persian, and Quechua), the average accuracy of MED+PT for 
SET1 is $0.9564$, compared to an overall average of $0.5808$. This observation shows clearly that
paradigm transduction obtains strong results if many 
forms per paradigm are given.

\paragraph{Effect of SHIP.}  
Further, \tabref{small_results} shows that 
SIG17+SHIP is better than
SIG17  by .0959 (.5971-.5012) on SET1,
.0779 (.7355-.6576) on SET2, and
.0301 (.8008-.7707) on SET3.
Stronger effects for smaller amounts of training
data indicate that SHIP's strategy of selecting a single reliable source 
is more important for weaker final models; in these cases,
selecting the most deterministic source reduces
errors due to noise. 

\katha{Include here results by languages.}

In contrast, the  performance of MED, the
neural model, is relatively independent of
the choice of source; this is in line with earlier
findings \cite{cotterell-sigmorphon2016}.  However, even for
MED+PT, adding SHIP (i.e., MED+PT+SHIP) slightly increases
accuracy  by .0061
(.7547-.7486) on SET2, and .0029
(.8483-.8454) on SET3 (L53).

\paragraph{Ablation.} 
MED does not perform well for either SET1 or SET2.
In contrast,  on SET3 it 
even outperforms SIG17
for a few languages.  However, MED loses against
MED+PT in all cases, highlighting the positive
effect of paradigm transduction.

Looking at PT next, even though PT
does not have a zero accuracy for any setting or language,
it performs consistently worse than MED+PT.  For  SET3, PT is even
lower than MED on average, by .3436 (.4211-.0775). 
Note that, in contrast to the other methods, PT's performance is not dependent
on the size of the training set. 
The main determinant for PT's
performance
is the size of the input subset during transductive
inference.
If the input subset is large, PT can perform better
than MED, e.g., for Hindi and Urdu.
For Khaling
SET1, PT even outperforms both MED and SIG17.
However, in most cases,
PT does not perform well on its
own.

MED+PT outperforms both
MED and PT. 
This confirms our initial intuition: MED
and PT learn complementary information for paradigm
completion. The base model learns the
general structure of the language (i.e.,
correspondences between tags and inflections) while
paradigm transduction teaches the model which character
sequences are common in a specific test paradigm.

\section{Analysis}
\seclabel{analysis}
\begin{figure}
\centering
\includegraphics[width=\columnwidth]{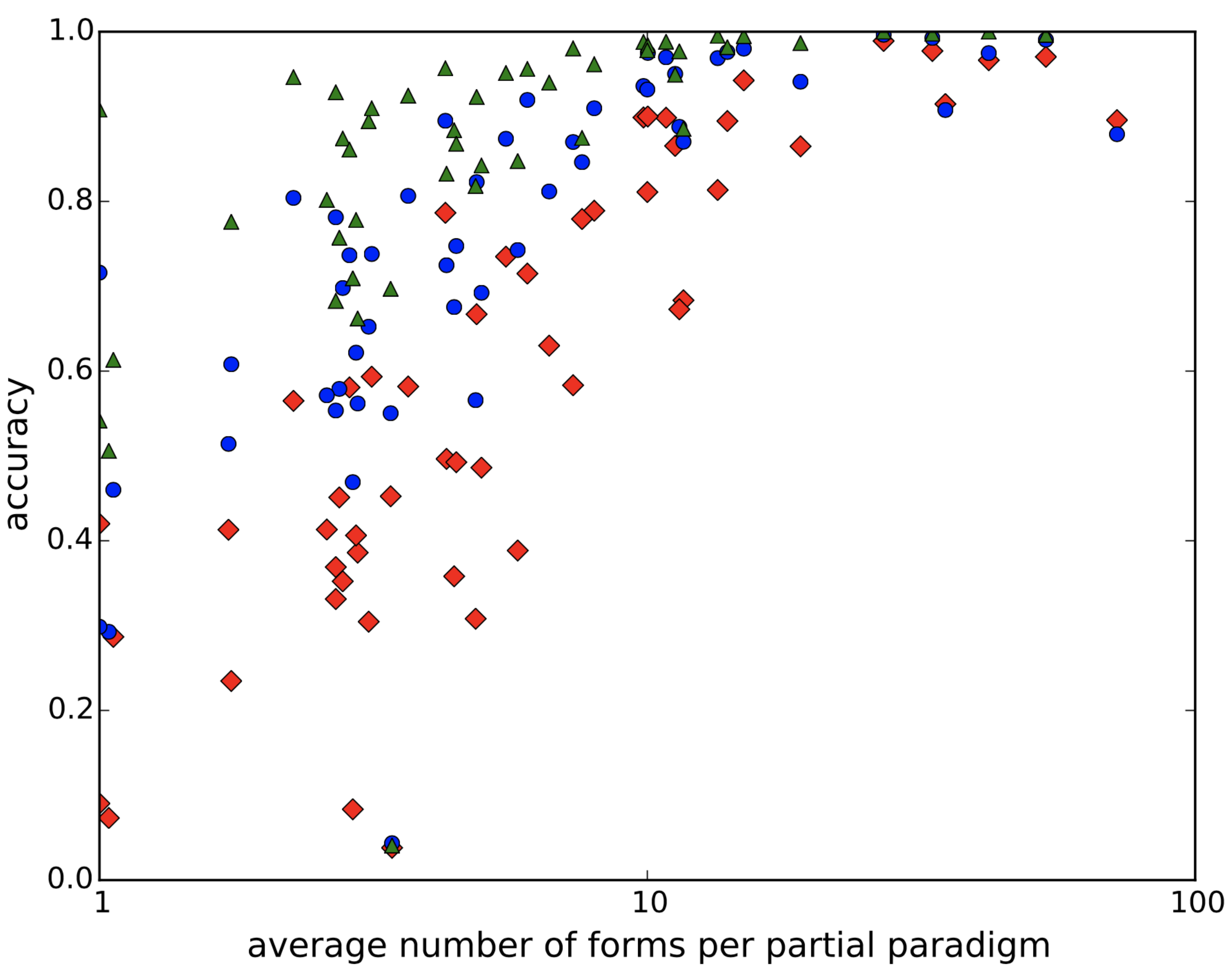}
\caption{Accuracy of MED+PT as a function of the average input
  subset size.
Red/diamonds: SET1; blue/circles: SET2; green/triangles: SET3.}
\figlabel{correlation}
\end{figure}
\subsection{On the Size of the Input Subset}
We expect paradigm transduction to become more effective as
the size of the input subset increases.
\figref{correlation} shows the accuracy of MED+PT as a
function of the average input subset size for SET1, SET2, and SET3.
Accuracy for languages
with input set sizes above 15
is higher than .8 in all settings.  In general,
languages with larger input set sizes perform better.
The correlation is not perfect because
languages have different degrees of morphological regularity.
However, the overall trend is clearly recognizable.

The
organizers of CoNLL--SIGMORPHON 
provided large input subsets in the development and test sets of languages
with large paradigms.
Thus, PT performs better for languages with
many inflected forms per paradigm, i.e., large $|T(w)|$.

\begin{table*}[t]
  \centering
  \setlength{\tabcolsep}{6pt}
  \begin{tabular}{l ||cccc|cccc|cccc}
    & \multicolumn{4}{c|}{SET1} & \multicolumn{4}{c|}{SET2} & \multicolumn{4}{c}{SET3} \\\hline
    & 1 & 2 & 4 & +PT & 1 & 2 & 4 & +PT & 1 & 2 & 4 & +PT \\\hline\hline
    dutch & .00 & .00 & .00 &  \textbf{.49} &  .04 & .01 & .00 & \textbf{.78} & .43 & .65 & .72 & \textbf{.87} \\
    german & .00 & .00 & .00 &  \textbf{.65} & .00 &  .00 & .01 &  \textbf{.75} & .44 & .42 & .59 & \textbf{.88} \\
    icelandic & .00 & .00 & .00 &  \textbf{.41} & .03 & .02 & .02 & \textbf{.50} & .24 & .33 & .35 & \textbf{.77} \\
    spanish & .00 & .00 & .00 &  \textbf{.92} & .03 & .09 & .09 & \textbf{.98} & .59 & .63 & .83 & \textbf{.99} \\
    welsh & .00 & .00 & .00 & \textbf{.91} & .05 & .14 & .15 & \textbf{.97} & .35 & .53 & .70 & \textbf{.99} 
  \end{tabular}
  \caption{MED accuracy
on five randomly selected languages
    with 1, 2, and 4 sources and
    combined with paradigm transduction (``+PT'').
Best results
in bold.\tablabel{multisource}}
\end{table*}
\subsection{On the Effect of Paradigm Transduction}
\seclabel{comparison}
We further analyze
why paradigm transduction improves the
performance of the base model MED, using the
German, English, and Spanish
SET2 examples for MED, PT, and MED+PT
given in \tabref{erroranalysis}.

\paragraph{German.} MED
generates an almost random sequence.
However, it learns that the umlaut ``\"{a}''
must appear in the target.  PT
only produces correct characters, but it produces far too many.
The reason may be that
the model is trained on both a double ``e''
and a double ``n'', learning that ``e'' and ``n'' are likely to
appear repeatedly.
MED+PT
generates the correct target.

\paragraph{English.} MED fails to generate ``h''  because
the bigram ``sh'' did not occur in training,
and so the
probability of ``h'' following ``s'' is estimated to be
low.  PT fails to produce
the suffix ``ing'', since it
does not occur in the input subset,
and, thus, PT has no way of learning it.
Again, MED+PT
generates the correct target.

\paragraph{Spanish.} MED produces ``crezcamos'',
a form that has the correct
tag V;SBJV;PRS;1;PL, but is a form of ``crecer'' (which
appears in the training set), not of ``creer'' (which does not).
This demonstrates  the problems resulting from a lack of lemma diversity during training. 
PT produces a combination of several of the forms in the
input subset:
subjunctive forms beginning with ``crey''
and ``creemos'' V;IND;PRS;1;PL. Again,
MED+PT
generates the correct target.

Overall, this analysis confirms that MED
learns relationships between paradigm cells,
while paradigm transduction adds knowledge about
the idiosyncracies of a partial test paradigm.

\subsection{Comparison to Multi-Source Models}
\label{subsec:multisource}
In this section, we explicitly compare our approach to 
neural multi-source models for morphological generation.

Following \newcite{kann-cotterell-schutze:2017:EACLlong}, we
employ attention-based RNN encoder-decoder networks  with two
or four input sources.  The input to a multi-source model is the
concatenation of all sources and corresponding tags. 
During training, we randomly sample (with repetition) one
or three additional forms from the paradigm of each example. 
At test time,
we sample the additional forms from the given partial paradigm;
without repetition
first, but repeating if not enough inflected forms
are 
available.  For autoencoding examples in the training
data, we simply concatenate two or four copies of the
source and the autoencoding tag.
We randomly select five languages for this experiment.

\tabref{multisource} shows that, for SET3,
four sources (column header ``4'') are generally better than
two sources (``2''), which in turn are better than one source
(``1''); thus, as expected, making additional sources available in
training improves results. We attribute one exception
(German accuracy is .4391 for ``1'' and .4179 for ``2'')
to the noisiness of the problem---training sets in terms of
number of paradigms are relatively small, even for SET3.

The improvements we see for SET3 are large.
This suggests that using more than four sources would
further improve results and perhaps reach the level of
performance of MED+PT, at the cost of a long training time.
However, for SET1 and SET2, there is no consistent
improvement from 1 to 2 to 4 sources.
While it is possible that
further optimization 
 could  improve the best multi-source result
given in \tabref{multisource}, the gap to MED+PT is very
large, and the improvement from 2 to 4 is small.
This indicates that multi-source methods cannot compete with
transductive learning for SET1 and SET2.

\begin{figure}[!htb]
\centering
\includegraphics[height=.97\textwidth,angle=0]{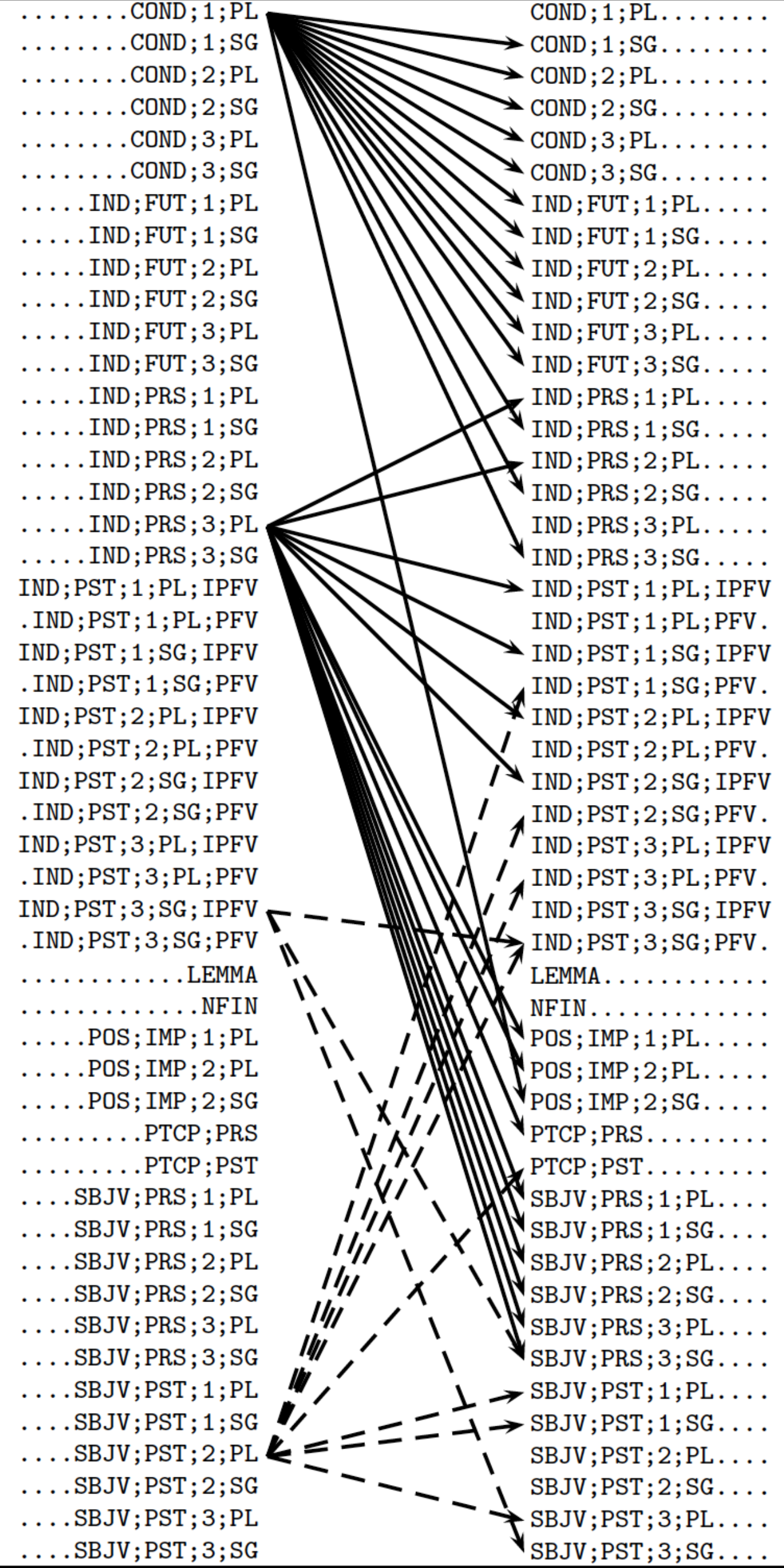}
\caption{\label{frenchship}Right: output set target to be
  generated. Left: input set source selected by SHIP. Arrows
  for the two most frequently selected sources are solid,
  arrows for the two least frequently selected sources are dashed.}
\end{figure}
\subsection{Qualitative Analysis of SHIP}
For a qualitative analysis of SHIP, we look at the sources it selects 
for French verbs on the development set; the complete diagram 
is shown in Figure \ref{frenchship}.
For most verbs, future and conditional  can be
predicted from
COND;1;PL
(e.g., ``finirions''), and indicative present, indicative
imparfait and subjunctive present  from IND;PRS;3;PL (e.g., ``finissent''). In
case of ties, SHIP selects the alphabetically first tag; this
explains why 
COND;1;PL
gets preference over
IND;PRS;3;PL for indicative present singular.
These two forms represent two of the \emph{principal parts} of
French conjugation, the infinitive (almost always derivable
from COND;1;PL) and the stem that is used for plural
indicative, imparfait, and other paradigm cells---which is
sometimes not derivable from the infinitive as is the case
for ``finir''.
In comparison,
IND;PST;3;SG;IPFV and SBJV;PST;2;PL are less reliable
sources. But they are still  reasonably accurate if no
better alternative is available; consider the following 
SBJV;PST;2;PL $\rightarrow$ IND;PST;1;SG;PFV generations: 
``parlassiez'' $\mapsto$ ``parlai'',
``finissiez'' $\mapsto$ ``finis'',
``missiez'' $\mapsto$ ``mis'',
``prissiez'' $\mapsto$ ``pris''.

We thus conclude that SHIP indeed learns to select appropriate source forms.

\section{Related Work}
\seclabel{rel_work}

\paragraph{Morphological generation.}
In the last two years, most work on  paradigm completion has been done in the context of the SIGMORPHON 2016 and the CoNLL--SIGMORPHON 2017 shared tasks
\cite{cotterell-sigmorphon2016,cotterell-conll-sigmorphon2017}.
Due to the success of neural seq2seq models in 2016 \cite{kann-schutze:2016:P16-2,aharoni-goldberg-belinkov:2016:SIGMORPHON}, 
systems developed for the 2017 edition were mostly neural
\cite{makarov-ST2017,bergmanis-kann-ST2017,zhou2017multi}.
Besides the shared task systems, \newcite{kann-ST2017} presented a paradigm completion model for a multi-source setting that made use of
an attention mechanism to decide which input form to attend to at each time step.
They used 
randomly chosen, independent pairs of source and target
forms for training. This differs crucially from the setting we consider
in that no complete paradigms were available in
their training sets.
Only \newcite{cotterell-sylakglassman-kirov:2017:EACLshort}
 addressed essentially the same task we do, but
they only considered
the high-resource setting:
their models were trained on hundreds of complete paradigms.
The experiments reported in \S\ref{subsec:multisource}
empirically confirm that inductive-only models perform
poorly in our setting.

Several
ways to employ neural models for morphological generation with limited data have been
proposed, 
e.g., semi-supervised training
\cite{zhou2017multi,kann2017unlabeled} or simultaneous
training on multiple languages \cite{kann2017one}.
The total number of sources in the training set
in some of our settings
may be
comparable to this earlier work, but
our training sets are less diverse since many forms come from
the same paradigm. We argue in \secref{intro} that the
number of paradigms (not the number of sources) 
measures the effective size of the training set.

Other important work on morphological generation---neural and non-neural---includes
\newcite{dreyer2008latent,durrett2013supervised,hulden-forsberg-ahlberg:2014:EACL,nicolai2015inflection,faruqui-EtAl:2016:N16-1,yin2015abcnn}.

\paragraph{Seq2seq models in NLP.}
Even though neural seq2seq models were originally designed for
machine translation \cite{sutskever2014sequence,cho2014properties,bahdanau2015neural}, their application 
has not stayed limited to this area.
Similar architectures have been successfully applied to many seq2seq tasks in NLP, e.g., syntactic
parsing \cite{vinyals2015grammar}, language correction \cite{xie2016neural}, normalization of historical texts \cite{P17-1031}, 
or text simplification \cite{textsimp}.
Transductive inference is similar to
domain adaptation, e.g., in
machine translation \cite{luong2015stanford}. One difference
is that  training set  and test set can hardly be called
different domains in paradigm completion. Another difference
is that explicit structured labels
(the morphological tags of the
forms in the input subset)
are available at test
time in paradigm completion.

\section{Conclusion}
We presented two new methods for minimal-resource paradigm
completion: paradigm transduction and 
SHIP.
 Paradigm transduction learns
general inflection rules through standard inductive training and
idiosyncracies of a test paradigm through transduction.
We showed that paradigm transduction effectively mitigates the problem of
overfitting due to a lack of
diversity in the training data.
SHIP is a robust non-neural method that identifies a single reliable
source for generating a target. In the minimal-resource
setting, this is an effective alternative to
learning how
to combine evidence from multiple sources.
Considering the average over all languages of a 52-language
benchmark dataset, we outperform the previous state 
of the art by at least $7.07\%$, and up to
$9.71\%$ absolute accuracy.

\section*{Acknowledgments}
We would like to thank Samuel Bowman, Ryan Cotterell, Nikita Nangia, and Alex Warstadt for their feedback on this work. This research was supported by the European Research Council (ERC \#740516).

\bibliography{emnlp2018}
\bibliographystyle{acl_natbib_nourl}

\appendix
\newpage
\onecolumn
\section{Detailed Experimental Results by Language}
\label{app:C}
\begin{table*}[h]
  \centering
  \scriptsize 
    \setlength{\tabcolsep}{2.25pt} 
  \begin{tabular}{r | l || c c c c c c | c c c c c c c | c c c c c c c}
                      && \multicolumn{6}{c|}{\bf SET1} & \multicolumn{7}{c|}{\bf SET2} & \multicolumn{7}{c}{\bf SET3} \\\hline
                          && \rotatebox{90}{COPY} &  \rotatebox{90}{PT} & \rotatebox{90}{SIG17} & \rotatebox{90}{SIG17+SHIP} & \rotatebox{90}{MED+PT} & \rotatebox{90}{MED+PT+SHIP} & \rotatebox{90}{COPY} & \rotatebox{90}{MED} & \rotatebox{90}{PT} & \rotatebox{90}{SIG17} & \rotatebox{90}{SIG17+SHIP} & \rotatebox{90}{MED+PT} & \rotatebox{90}{MED+PT+SHIP} & \rotatebox{90}{COPY} & \rotatebox{90}{MED} & \rotatebox{90}{PT} & \rotatebox{90}{SIG17} & \rotatebox{90}{SIG17+SHIP} & \rotatebox{90}{MED+PT} & \rotatebox{90}{MED+PT+SHIP} \\\hline\hline
1 & albanian & .0385
& .0305 & .1224 & .2742 & \textbf{.6793} & .6568 & .0385 & .0055 & .0305 & .8362 & .2335 & .8786 & \textbf{.8891} & .0385 & .1808 & .0305 & .8951 & .4276 & \textbf{.9750} & .9700 \\
2 & arabic & .0334
& .0445 & .4262 & .4744 & \textbf{.8191} & .8006 & .0334 & .0378 & .0445 & .5389 & .7265 & .8962 & \textbf{.9096} & .0334 & .3988 & .0445 & .5567 & .7509 & .9400 & \textbf{.9526} \\
3 & armenian & .0233
& .0841 & .7912 & .9002 & \textbf{.9235} & .9174 & .0233 & .0319 & .0841 & .8089 & .9706 & .9762 & \textbf{.9787} & .0233 & .5297 & .0841 & .8611 & .8049 & \textbf{.9919} & .9899 \\
4 & basque
& .0004
& .6743 & .0050 & .1034 & .9298 & \textbf{.9427}
& .0004 
& .0108 & .6743 & .0457 & .0968 & .9132 & \textbf{.9190} & - & - & - & - & - & - & - \\
5 & bengali & .0357
& .0127 & .7720 & \textbf{.9019} & .6675 & .6599 & .0357 & .0102 & .0127 & .8586 & \textbf{.9172} & .8344 & .8089 & .0357 & .0701 & .0127 & .8752 & \textbf{.9248} & .8675 & .8713 \\
6 & bookmal & .1321
& .0377 & .4025 & \textbf{.4151} & .1069 & .1069 & .1321 & .0377 & .0377 & \textbf{.5094} & \textbf{.5094} & .3585 & .3585 & .1321 & .4277 & .0377 & .6792 & \textbf{.6855} & .5094 & .5157 \\
7 & bulgarian & .0854
& .0151 & .3920 & \textbf{.4958} & .4807 & .4774 & .0854 & .0318 & .0151 & .4908 & \textbf{.5444} & .5243 & .5427 & .0854 & .2663 & .0151 & .7471 & \textbf{.8425} & .8208 & .8191 \\
8 & catalan & .0173
& .0210 & \textbf{.9407} & .9346 & .8758 & .8655 & .0173 & .0415 & .0210 & .9533 & \textbf{.9706} & .9659 & .9641 & .0173 & .5626 & .0210 & .9603 & .9795 & \textbf{.9897} & .9874 \\
9 & czech & .1255
& .0498 & .2656 & \textbf{.3503} & .2873 & .3122 & .1255 & .0166 & .0498 & .5643 & \textbf{.5873} & .5332 & .5373 & .1255 & .2739 & .0498 & .8579 & \textbf{.8794} & .8444 & .8423 \\
10 & danish & .1738
& .0230 & \textbf{.4230} & \textbf{.4230} & .1705 & .1738 & .1738 & .0492 & .0230 & \textbf{.7016} & \textbf{.7016} & .4918 & .5016 & .1738 & .4984 & .0230 & \textbf{.7574} & .7541 & .3541 & .4066 \\
11 & dutch & .1661
& .0018 & .4963 & \textbf{.5664} & .4908 & .4982 & .1661 & .0424 & .0018 & .6771 & \textbf{.8653} & .7841 & .7915 & .1661 & .4317 & .0018 & .7804 & \textbf{.8856} & .8727 & .8727 \\
12 & english & .2080
& .0240 & \textbf{.7640} & \textbf{.7640} & .4000 & .4000 & .2080 & .0600 & .0240 & \textbf{.8400} & \textbf{.8400} & .7360 & .7360 & .2080 & .6680 & .0240 & \textbf{.9120} & \textbf{.9120} & .8440 & .8440 \\
13 & estonian & .0373
& .0210 & .3888 & .6857 & .7881 & \textbf{.7992} & .0373 & .0081 & .0210 & .6257 & .7119 & \textbf{.9173} & \textbf{.9173} & .0373 & .2893 & .0210 & .7730 & .7189 & .9761 & \textbf{.9796} \\
14 & faroese & .1570
& .0239 & .4978 & \textbf{.5755} & .3991 & .3946 & .1570 & .0149 & .0239 & .5919 & \textbf{.6831} & .6069 & .6024 & .1570 & .3602 & .0239 & .7010 & \textbf{.7549} & .7175 & .7160 \\
15 & finnish & .0650
& .0257 & .6082 & \textbf{.7630} & .6330 & .6236 & .0650 & .0274 & .0257 & .6279 & \textbf{.8896} & .8777 & .8760 & .0650 & .3311 & .0257 & .6929 & .9264 & \textbf{.9615} & .9504 \\
16 & french & .0213
& .0142 & .8709 & \textbf{.9040} & .8918 & .8943 & .0213 & .0930 & .0142 & .8516 & .9233 & .9487 & \textbf{.9517} & .0213 & .6880 & .0142 & .9263 & .9548 & .9964 & \textbf{.9980} \\
17 & gaelic & .3684
& .0202 & .4413 & \textbf{.5061} & .4170 & .3927 &.3684
& .0000 
& .0202 & .4130 & \textbf{.4858} & .4413 & .4575 & - & - & - & - & - & - & - \\
18 & georgian & .0534
& .0036 & .7886 & \textbf{.8682} & .7684 & .7613 & .0534 & .1235 & .0036 & .8242 & \textbf{.8967} & .8753 & .8646 & .0534 & .6211 & .0036 & .9097 & \textbf{.9549} & .9335 & .9477 \\
19 & german & .2959
& .0174 & .6983 & \textbf{.7466} & .6460 & .6267 & .2959 & .0000 
& .0174 & .7041 & \textbf{.8375} & .7485 & .7563 & .2959 & .4391 & .0174 & .7640 & .8549 & \textbf{.8743} & \textbf{.8743} \\
20 & haida & .0087
& .0981 & .4715 & .9452 & .9279 & \textbf{.9611} &
.0087
& .0000 
& .0981 & .6453 & .9481 & .9394 & \textbf{.9762} & - & - & - & - & - & - & - \\
21 & hebrew & .0507
& .0178 & .3238 & .6148 & \textbf{.6548} & .6459 & .0507 & .2109 & .0178 & .4270 & .6530 & \textbf{.8639} & .8585 & .0507 & .5329 & .0178 & .5427 & .6984 & .9137 & \textbf{.9377} \\
22 & hindi & .0038
& .6005 & .6449 & .7082 & \textbf{.9405} & .9264 & .0038 & .0000 
& .6005 & .7031 & .8565 & \textbf{.9504} & .9496 & .0038 & .3621 & .6005 & .9539 & .4922 & .9979 & \textbf{1.000} \\
23 & hungarian & .0310
& .0038 & .1791 & \textbf{.5489} & .5435 & .5488 & .0310 & .0060 & .0038 & .4452 & \textbf{.7621} & .7498 & .7415 & .0310 & .3447 & .0038 & .5412 & .8237 & .9108 & \textbf{.9146} \\
24 & icelandic & .1640
& .0148 & .4579 & \textbf{.5657} & .4106 & .4136 & .1640 & .0295 & .0148 & .5451 & \textbf{.6721} & .4993 & .5288 & .1640 & .2393 & .0148 & .6691 & .7474 & \textbf{.7696} & .7430 \\
25 & irish & .1971
& .0128 & .3230 & \textbf{.4343} & .3248 & .3285 & .1971 & .0109 & .0128 & .4033 & .4872 & .5675 & \textbf{.5766} & .1971 & .2518 & .0128 & .4799 & .5712 & \textbf{.6788} & .6588 \\
26 & italian & .0186
& .0100 & .6643 & \textbf{.7200} & .6962 & .6976 & .0186 & .0190 & .0100 & .7186 & .7767 & .9052 & \textbf{.9233} & .0186 & .3171 & .0100 & .7305 & .7848 & \textbf{.9733} & .9719 \\
27 & khaling
& .0209 & .5773 & .4230 & .4422 & .9686
& \textbf{.9716}
& .0007 
& .2258 & .5773 & .5820 & .6208 & \textbf{.9918} &
.9895
& .0007 
& .6979 & .5773 & .7908 & .6583 & .9960 & \textbf{.9983} \\
28 & kurmanji & .0817
& .0174 & .7843 & \textbf{.8017} & .4591 & .4539 & .0817 & .0765 & .0174 & .8835 & \textbf{.8887} & .7965 & .8330 & .0817 & .7165 & .0174 & \textbf{.9287} & .9165 & \textbf{.9287} & .9183 \\
29 & latin & .0600
& .0146 & .2416 & \textbf{.5198} & .3016 & .2899 &
.0600
& .0000 
& .0146 & .3909 & \textbf{.8507} & .4905 & .5168 & .0600 & .2050 & .0146 & .4671 & \textbf{.8814} & .6852 & .7379 \\
30 & latvian & .0735
& .0115 & .6888 & \textbf{.7579} & .4971 & .5115 & .0735 & .0101 & .0115 & .7983 & \textbf{.8919} & .7752 & .7680 & .0735 & .5086 & .0115 & .8646 & \textbf{.9669} & .9179 & .9150 \\
31 & lithuanian & .0169
& .0125 & .3845 & \textbf{.4951} & .4790 & .4835 & .0169 & .0134 & .0125 & .6619 & \textbf{.8287} & .7395 & .7431 & .0169 & .2810 & .0125 & .6048 & \textbf{.8582} & .8260 & .8359 \\
32 & maced. & .0959
& .0072 & .4249 & \textbf{.6023} & .4149 & .4077 & .0959 & .0615 & .0072 & .8655 & \textbf{.8898} & .7611 & .7954 & .0959 & .5722 & .0072 & .8970 & \textbf{.9714} & .9614 & .9599 \\
33 & navajo & .0589
& .0014 & .2658 & \textbf{.3563} & .0466 & .0370 & .0589 &
.0616 & .0014 & .3315 & \textbf{.4801} & .0329 & .0315 &
.0589 & .2849 & .0014 & .3795 & \textbf{.5695} & .0301 &
.0137 \\
34 & nynorsk & .1043
& .0245 & \textbf{.4233} & \textbf{.4233} & .1043 & .1043 & .1043 & .0123 & .0245 & \textbf{.6074} & \textbf{.6074} & .3067 & .3129 & .1043 & .4785 & .0245 & .6442 & \textbf{.6626} & .4172 & .4294 \\
35 & persian & .0218
& .2400 & .7342 & .1633 & \textbf{.9880} & .9844 & .0218 & .0509 & .2400 & .7829 & .1815 & .9898 & \textbf{.9960} & .0218 & .6073 & .2400 & .7644 & .1971 & \textbf{.9993} & .9989 \\
36 & polish & .0986
& .0218 & .5672 & \textbf{.6453} & .5928 & .5826 & .0986 & .0205 & .0218 & .8028 & \textbf{.8271} & .7631 & .7452 & .0986 & .3995 & .0218 & \textbf{.9001} & .8924 & .8515 & .8553 \\
37 & portug. & .0697
& .0432 & .9171 & \textbf{.9694} & .9135 & .9190 & .0697 & .1058 & .0432 & .9529 & \textbf{.9858} & .9784 & .9742 & .0697 & .6829 & .0432 & .9616 & \textbf{.9861} & .9839 & .9823 \\
38 & quechua & .0083
& .5093 & .9133 & \textbf{.9998} & .9835 & .9856 & .0083 & .2750 & .5093 & .9134 & \textbf{.9999} & .9887 & .9926 & .0083 & .7852 & .5093 & .8913 & \textbf{.9994} & .9965 & \textbf{.9994} \\
39 & romanian & .0754
& .0133 & .1420 & \textbf{.2500} & .1317 & .1213 & .0754 & .0074 & .0133 & .6154 & \textbf{.7663} & .5814 & .6124 & .0754 & .2914 & .0133 & \textbf{.7988} & .7825 & .7500 & .7544 \\
40 & russian & .1166
& .0092 & .4018 & \textbf{.4617} & .3221 & .3528 & .1166 & .0598 & .0092 & .8298 & \textbf{.8574} & .7715 & .7761 & .1166 & .6610 & .0092 & .8589 & \textbf{.8712} & .8635 & .8650 \\
41 & sami & .0436
& .0048 & .1562 & \textbf{.3986} & .2890 & .3046 & .0436 & .0155 & .0048 & .3172 & \textbf{.8351} & .7856 & .7818 & .0436 & .2250 & .0048 & .4568 & .9108 & .9108 & \textbf{.9176} \\
42 & SCB & .0870
& .2802 & .3007 & .3116 & \textbf{.5266} & .5133 &.0870
& .0000 
& .2802 & .3684 & .3744 & \textbf{.6401} & .6196 & .0870 & .0918 & .2802 & .8575 & .7536 & .8829 & \textbf{.8841} \\
43 & slovak & .1301
& .0053 & .4439 & \textbf{.5365} & .3226 & .3012 & .1301 & .0160 & .0053 & .5936 & \textbf{.6667} & .4011 & .3993 & .1301 & .3119 & .0053 & .6934 & \textbf{.7184} & .6310 & .6275 \\
44 & slovene & .1094
& .0151 & .5774 & \textbf{.7928} & .5925 & .5867 & .1094 & .0012 & .0151 & .6787 & .8510 & .8254 & \textbf{.8661} & .1094 & .4505 & .0151 & .7648 & .9302 & \textbf{.9430} & .9336 \\
45 & sorani & .0110
& .1285 & .5478 & .5435 & .6162 & \textbf{.6272} & .0110 & .0042 & .1285 & .6839 & .5773 & .8216 & \textbf{.8225} & .0110 & .2984 & .1285 & .7210 & .6044 & .8512 & \textbf{.8529} \\
46 & sorbian & .0799
& .0237 & .3758 & .5643 & .5793 & \textbf{.6080} & .0799 & .1024 & .0237 & .6592 & \textbf{.8402} & .7328 & .7553 & .0799 & .5031 & .0237 & .8227 & \textbf{.8901} & .8664 & .8739 \\
47 & spanish & .0144
& .0207 & .8067 & .8835 & \textbf{.9186} & .9088 & .0144 & .0330 & .0207 & .9218 & .8884 & \textbf{.9768} & .9709 & .0144 & .5884 & .0207 & .9358 & .8891 & .9853 & \textbf{.9874} \\
48 & swedish & .1824
& .0441 & .4353 & \textbf{.5118} & .2735 & .2735 & .1824 & .0441 & .0441 & .5735 & \textbf{.7088} & .5176 & .5265 & .1824 & .4206 & .0441 & .7824 & \textbf{.8471} & .7235 & .7353 \\
49 & turkish & .0168
& .0894 & .2050 & .6984 & \textbf{.8748} & .8643 & .0168 & .0358 & .0894 & .7204 & .9258 & \textbf{.9754} & .9740 & .0168 & .5388 & .0894 & .8673 & .9402 & .9934 & \textbf{.9941} \\
50 & ukrainian & .0873
& .0159 & .4222 & \textbf{.4921} & .3460 & .3206 & .0873 & .0413 & .0159 & \textbf{.6714} & .6619 & .5857 & .5984 & .0873 & .3984 & .0159 & .7397 & \textbf{.7619} & .7413 & .7508 \\
51 & urdu & .0264
& .5434 & .8059 & .8059 & \textbf{.8754} & .8720 & .0264 & .0014 & .5434 & .8102 & .8102 & .9438 & \textbf{.9485} & .0264 & .2007 & .5434 & .9533 & .7040 & .9887 & \textbf{.9901} \\
52 & welsh & .0277
& .0147 & .5167 & .4319 & .9104 & \textbf{.9128} & .0277 & .0513 & .0147 & .8280 & .8843 & .9707 & \textbf{.9756} & .0277 & .3504 & .0147 & .8525 & .9446 & .9886 & \textbf{.9919} \\\hline
53 & {\bf average} & .0810
& .0883 & .5012 & \textbf{.5971} & .5808 & .5793 & .0810 & .0432 & .0883 & .6576 & .7355 & .7486 & \textbf{.7547} & .0782 & .4211 & .0775 & .7707 & .8008 & .8454 & \textbf{.8483} 

  \end{tabular}
  \caption{Accuracy on PC 
  for SIG17+SHIP
  (the shared task baseline SIG17 with  SHIP), MED+PT (MED with paradigm transduction),
  and MED+PT+SHIP (MED with paradigm transduction and
  SHIP). We omit the column ``MED'' for SET1: all results
  were $<.001$ except for .0123 for Navajo.
bookmal = Norwegian Bokm\aa l,
gaelic = Scottish Gaelic,
maced.\ = Macedonian,
nynorsk = Norwegian Nynorsk,
portug.\ = Portuguese,
  sami = Northern Sami,
  SCB =
  Serbo-Croatian,
  sorbian = Lower Sorbian. Since Basque, Scottish Gaelic, and Haida are low-resource languages, no SET3 data is available. 
  \tablabel{results}}
\end{table*}

\end{document}